\documentclass[10pt,twocolumn,letterpaper]{article}

\usepackage{cvpr}              

\usepackage{multirow}
\usepackage{array}
\usepackage{colortbl}
\colorlet{light-gray}{black!20}

\newcolumntype{L}[1]{>{\raggedright\let\newline\\\arraybackslash\hspace{0pt}}m{#1}}
\newcolumntype{C}[1]{>{\centering\let\newline\\\arraybackslash\hspace{0pt}}m{#1}}
\newcolumntype{R}[1]{>{\raggedleft\let\newline\\\arraybackslash\hspace{0pt}}m{#1}}

\newcommand{\parens}[1]{\left(#1\right)}
\newcommand{\braces}[1]{\left\{#1\right\}}
\newcommand{\bracks}[1]{\left[#1\right]}
\newcommand{\modulus}[1]{\left\vert#1\right\vert}
\newcommand{\norm}[1]{\left\Vert#1\right\Vert}

\newcommand{\mpage}[2]
{
\begin{minipage}{#1\linewidth}\centering
#2
\end{minipage}
}

\makeatother

\definecolor{AGcolor}{rgb}{0.8, 0.0, 0.8}

\definecolor{UBcolor}{rgb}{0.0, 0.8, 0.6}

\def\ours{SimMotionEdit\xspace} 

\definecolor{cvprblue}{rgb}{0.21,0.49,0.74}
\usepackage[pagebackref,breaklinks,colorlinks,allcolors=cvprblue]{hyperref}

\def\paperID{5802} 
\def\confName{CVPR}
\def\confYear{2025}

\title{SimMotionEdit: Text-Based Human Motion Editing with Motion Similarity Prediction}

\author{
  Zhengyuan Li\textsuperscript{1}\quad
  Kai Cheng\textsuperscript{1}\quad
  Anindita Ghosh\textsuperscript{2}\quad
  Uttaran Bhattacharya\textsuperscript{3}\quad
  Liangyan Gui\textsuperscript{4}\quad
  Aniket Bera\textsuperscript{1}\vspace{6pt}\\
\textsuperscript{1}Purdue University\quad
\textsuperscript{2}DFKI, MPI-INF, Saarland Informatics Campus\\
\textsuperscript{3}Adobe Inc.\quad
\textsuperscript{4}University of Illinois Urbana-Champaign\vspace{3pt}}

\begin{document}
\maketitle
\begin{abstract}
    Text-based 3D human motion editing is a critical yet challenging task in computer vision and graphics. While training-free approaches have been explored, the recent release of the MotionFix dataset, which includes source-text-motion triplets, has opened new avenues for training, yielding promising results. However, existing methods struggle with precise control, often leading to misalignment between motion semantics and language instructions. In this paper, we introduce a related task --- motion similarity prediction --- and propose a multi-task training paradigm, where we train the model jointly on motion editing and motion similarity prediction to foster the learning of semantically meaningful representations. To complement this task, we design an advanced Diffusion-Transformer-based architecture that separately handles motion similarity prediction and motion editing. Extensive experiments demonstrate the state-of-the-art performance of our approach in both editing alignment and fidelity. Project URL: \url{https://github.com/lzhyu/SimMotionEdit}.
\end{abstract}
    
\section{Introduction}
\label{sec:intro}

Human motion synthesis plays a central role in diverse applications, including character animation~\cite{starke2020local, s2ag, ghosh2023imos}, human-robot collaborations~\cite{koppula2013anticipating, andrist2015look}, and autonomous driving~\cite{paden2016survey}. Current approaches primarily focus on initial control signals, such as manipulating joint positions~\cite{omnicontrol, tlcontrol} or applying fine-grained language instructions~\cite{jin2024act, zhang2024finemogen}. With advancements in text-based motion generation~\cite{t2g, ghosh_textbased2021, tm2t, mdm, guo2024momask}, the ability to control and edit human motions has emerged as the next frontier, which can enable animators to refine motion sequences in animation workflows to fit specific requirements.
In this work, we aim to address this need by focusing on editing existing human motions rather than generating them solely from other conditioning inputs.

Recent works~\cite{raab2024monkey, chen2024motionclr} have leveraged pre-trained motion diffusion models with attention manipulation to edit motion features, achieving effects such as style transfer and word emphasis. However, these methods are limited in scope due to the lack of dedicated model training and supervision. With the release of the text-based human motion editing dataset MotionFix~\cite{athanasiou2024motionfix}, we can now incorporate robust supervision to develop \ours, a system that enables free-form text-based human motion editing.

\begin{figure}[t]
  \centering
   \includegraphics[width=\linewidth]{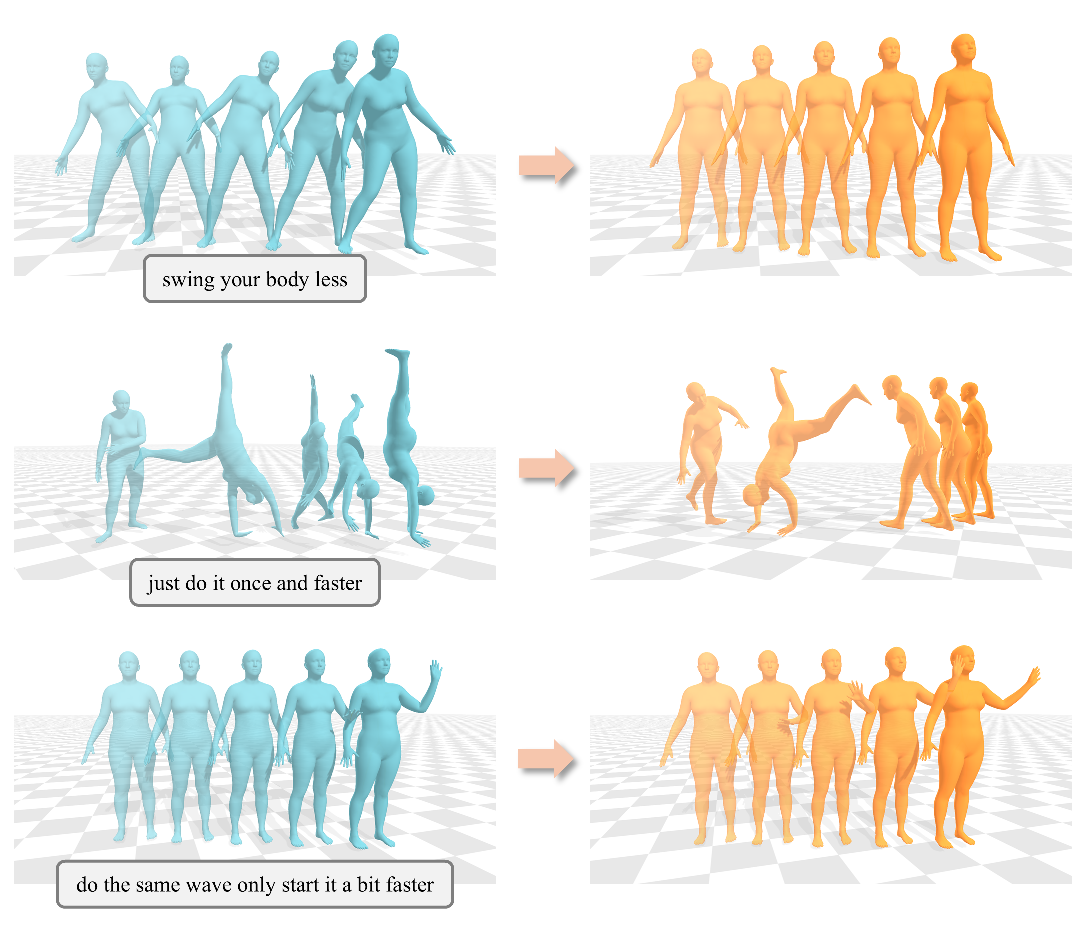}
   \caption{\textbf{Text-Based Motion Editing.} Our method \ours generates edited human motion sequences from text instructions and source motion sequences. }
   \label{fig:teaser}
\end{figure}

The primary challenge in this task lies in aligning the generated motion with both the source motion and the text instructions, as they jointly define the desired outcome. Enhancing the interaction between the source motion and the text instructions is thus crucial for improved performance.
To this end, the success of multi-task learning \cite{zhang2021survey} has demonstrated that models can achieve better generalization by leveraging knowledge from other tasks. Auxiliary loss, a key concept in multi-task learning, has proven effective across various domains, including image classification~\cite{szegedy2015going}, scene parsing~\cite{zhao2017pyramid}, reinforcement learning~\cite{lample2017playing}, and code generation~\cite{ding2024horizon}. The essential criterion in designing auxiliary tasks is that they should support the learning of representations that benefit the main task.
Consider the example of an animator instructed to edit a human motion sequence based on an editing description and a source motion. The animator does not begin editing immediately --- a common first step is to identify the keyframes to modify. Similarly, for our model, identifying keyframes to edit is not only essential but also beneficial to performance. In many cases, the keyframes to edit are predictable. For example, given the instruction ``don't turn back at the end; instead, keep walking'', it is reasonable to anticipate frame changes at the point where the character starts turning back. As a result, the frames where the edit happens are dissimilar to the corresponding frames in the source motion. Following this intuition, we apply techniques such as sliding windows, value normalization, and weight balancing across different similarity measures to construct a predictable time series of motion similarity. We formulate the task as \textit{motion similarity prediction}, where the model predicts the similarity between the source and the edited motions given the edit instructions.

Given the multitask nature of learning motion editing via motion similarity prediction, we opt for the diffusion framework due to its flexibility. While various network architectures~\cite{mdm, zhang2024finemogen, zhang2024large} have been proposed for human motion synthesis, they are not directly applicable to human motion editing, as most of them are designed to operate solely with text as input. Further, we separate our neural network into two parts: a condition transformer for motion similarity prediction and a diffusion transformer~\cite{peebles2023scalable} for motion editing. This design choice offers two key advantages. First, the conditional transformer helps enhance the text and source motion features to embed information on motion similarity. The diffusion transformer effectively leverages those embeddings to generate the edited motion. Second, the condition transformer handles motion similarity prediction \textit{decoupled} from the diffusion process, which allows each part of the neural network to specialize in its individual task and expedites as well as stabilizes training convergence.

We summarize our key contributions as follows:
\begin{itemize}
    \item {
        \emph{\ours}, a text-based human motion editing framework with an auxiliary task of motion similarity prediction to efficiently inter-relate source motion features and instruction text features.
    }
    
    \item {
        A Motion Diffusion Transformer architecture, trained on motion similarity prediction and edited motion generation to effectively learn text-based motion editing.
    }

    \item {
        Demonstrating the state-of-the-art performance of our architecture and our auxiliary task through extensive quantitative and qualitative experiments on the open-source dataset of MotionFix~\cite{athanasiou2024motionfix}.
    }
\end{itemize}

\section{Related Works}
\label{sec:rw}
We briefly discuss existing literature on the related areas of 3D human motion editing, diffusion models for motion generation, and auxiliary tasks used to train neural networks.

\paragraph{3D Human Motion Editing.}
Earlier works on 3D human motion editing have explored editing motions with space or time constraints~\cite{gleicher97, mopathedit, lee_edit}. However, generating realistic and coordinated motion editing typically requires users to manually provide such specific constraints.
To this end, deep learning approaches have explored automating the process of human motion editing with simple inputs by incorporating different sub-tasks, such as motion stylization~\cite{aberman2020unpaired, 10030800}, pose editing~\cite{oreshkin2022protores}, and in-betweening~\cite{edge, mdm, shafir2023human, twostagetransformer}. While techniques like motion stylization can perform global edits on motions, they do not enable precise control. Further, these approaches cannot be applied to the free-form text-based setting. 
\citet{kim2023flame} edit motions through diffusion inpainting, which requires a mask as input. \citet{raab2024monkey} perform motion following by replacing the queries in the self-attention. MotionCLR~\cite{chen2024motionclr} leverages attention manipulation to provide a set of motion editing methods. However, it does not directly take in texts as edit conditions.
\citet{goel2024iterative} heuristically select keyframes to edit certain types of motions. For text-based human pose editing, PoseFix~\cite{Delmas_2023_ICCV} enables free-form text as input. \citet{zhang2024finemogen} relies on large language models to rewrite the script and generate edited motions and only considers the original motion at the text level. However, it cannot be applied without providing a description of the source motion. \citet{CoMo_ECCV2024} prompts GPT to edit latent codes and generate edited human motions. MotionFix~\cite{athanasiou2024motionfix} provides a dataset and proposes a diffusion-based method to perform motion editing. Our task follows MotionFix~\cite{athanasiou2024motionfix}, and we propose a novel architecture and an auxiliary task to improve generative performance.
 
\begin{figure*}[t]
  \centering
   \includegraphics[width=\textwidth]{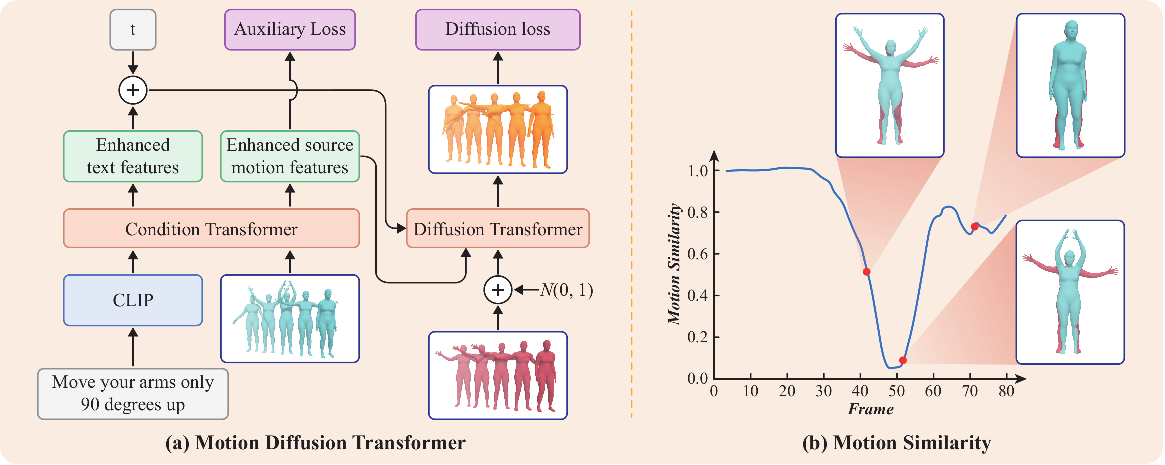}
   \caption{\textbf{Overview of \ours.} (a) The architecture consists of two modules: the condition transformer and the diffusion transformer. The condition transformer performs the auxiliary task of motion similarity prediction and enables the source motion features and the text features to mix. The diffusion transformer takes in the enhanced text features, the embedded diffusion step $t$ as the condition, the noisy edited motion, and the enriched source motion features, and predicts the denoised edited motions. (b) The auxiliary task motion similarity prediction is inspired by the fact that, given the text instructions, the similarity between source and edited motions is predictable. We use \textcolor{blue}{blue} for the source motion, \textcolor{red}{red} for the edited motion, and \textcolor{orange}{orange} for the generated motion.}
   \label{fig:pipeline}
\end{figure*}

\paragraph{Motion Generation via Diffusion Models.}
Diffusion models~\cite{ddpm1} have been widely applied to human motion generation from text~\cite{mdm, mofusion, zhang2022motiondiffuse, ren2023insactor, liang2024intergen,xu2023interdiff, remodiffuse, mld, motionlcm,liang2024omg} and other modalities~\cite{rempeluo2023tracepace, edge,Wei2023UnderstandingTM, ghosh2024remos, sui2025surveyhumaninteractionmotion}.The state-of-the-art motion diffusion models collapse text input into a single token. \citet{shafir2023human} modify MDM's~\cite{mdm} motion inpainting process to control end-effector trajectories but require dense multi-modal inputs like entire joint trajectories instead of the free-form text-based instructions.
With the introduction of recent text-motion 3D human datasets~\cite{lin2023motionx, babel, Guo_2022_CVPR},
there has been increased interest in conditioning the generation on free-form language inputs, in which \cite{zhang2022motiondiffuse, mdm, wan2023diffusionphase, shafir2023human, omnicontrol, chen2024motionclr} have
produced impressive results in free-from-text-based 3D human motion generation using diffusion models.

\paragraph{Auxiliary Tasks.}
Auxiliary tasks have played a significant role in enhancing the training of neural networks. Earlier works such as GoogLeNet (Inception v1)~\cite{szegedy2015going} added an auxiliary loss at intermediate layers to provide additional gradient signals to mitigate vanishing gradients. PSPNet\cite{zhao2017pyramid} incorporated an auxiliary loss at intermediate layers for the scene parsing task. Auxiliary losses have also been leveraged in techniques including deep reinforcement learning~\cite{lample2017playing,jaderberg2016reinforcement} and code generation~\cite{ding2024horizon} to enhance learning stability and feature extraction. Recently, REPA~\cite{yu2024representation} finds that aligning the features of the denoiser and pre-trained discriminative models substantially boosts the training of diffusion models. Other works~\cite{tarvainen2017mean,ruder2017overview} have also shown that auxiliary tasks can significantly improve the model's generalization by providing complementary signals to align the model's internal representations. However, to the best of our knowledge, auxiliary losses have not been explored previously for text-based human motion editing.
\section{Method}
\label{sec:formatting}

Given a text instruction $L$ and a source motion sequence $X = [x^{0}, x^{1},..., x^{F}]$ with $F$ frames, our objective is to generate an edited sequence $M = [m^{0}, m^{1},...,m^{F^\prime}]$ with $F^\prime$ frames, where each $m^{i}\in \mathbb{R}^{D}$ represents a human pose, $D$ being the feature dimension. Following MotionFix~\cite{athanasiou2024motionfix}, each $m^{i}$ consists of global velocity, global orientation, joint rotations, and local joint positions, totaling $D=207$.
We illustrate our network architecture, the Motion Diffusion Transformer, in \cref{fig:pipeline}. It consists of a condition transformer and a diffusion transformer. The condition transformer processes the source motion $X$ and the text instructions $L$, and provides enhanced condition features via motion similarity prediction (\cref{sec:msp}). The diffusion transformer takes in the concatenated sequence of enhanced source motion features and noised edited motion, and uses the enhanced text features as conditions, to generate edited motions (\cref{sec:dit}). We train the network jointly on motion similarity prediction and diffusion-based motion editing (\cref{sec:diffusion}).

\subsection{Diffusion-Based Motion Editing}
\label{sec:diffusion}
Denoising diffusion models~\cite{sohl2015deep, ddpm1} are generative models that use a structured sequence of transformations based on a Markov chain.
The diffusion process comprises a forward step, $q\parens{M_t \mid M_{t-1}}$, where we iteratively add Gaussian noise to a clean edited motion sample $M_0$ over $T$ timesteps until it resembles the distribution of $p\parens{M_T}$:
\begin{equation}
    M_t =\sqrt{\bar \alpha_{t}}M_0 + \sqrt{1-\bar\alpha_{t}} \epsilon,
\end{equation}
where $\bar \alpha_{t} \in \parens{0, 1}$ controls the rate of diffusion and $t \in \bracks{0, T}$.
When $\bar \alpha_{t}$ is small enough, we can approximate $M_T \approx\epsilon\sim \mathcal{N}\parens{\mathbf{0}, \mathbf{I}}$, $\mathbf{I}$ being the unit matrix.
In the subsequent reverse step, the diffusion process samples motions from a multivariate Gaussian distribution $M_T \sim \mathcal{N}\parens{\mathbf{0}, \mathbf{I}}$, and we obtain the conditional probability of $M_0$ as
\begin{equation}
    p_\theta\parens{M_0 \mid c} = \int_{M_1}^{M_T} p\parens{m_T} \prod_{u=1}^T p_\theta\parens{m_{u-1} \mid m_u, c} \mathrm{d}\mathbf{m}.
\end{equation}

For text-conditioned motion editing, we approximate $p_\theta(\parens{M_{t-1} \mid M_t}$ with a learnable function $\mathcal{E}\parens{M_t, t, L, X}$, which conditions the edited motion on both the text instruction $L$ and the source motion $X$.
Following recent approaches~\cite{ramesh2022hierarchical,mdm,athanasiou2024motionfix,goel2024iterative}, we predict the original motion signal $M_0 = \mathcal{E}\parens{M_t, t, L, X}$ via the editing loss function,
\begin{equation}
    \mathcal{L}_e = \mathbb{E}_{M_0 \sim q\parens{M_0 \mid L, X}, t \sim \bracks{1, T}} \bracks{\norm{M_0 - \mathcal{E}\parens{M_t, t, L, X}}_2^2}.
    \label{eqn:loss_editing}
\end{equation}

\subsection{Motion Similarity Prediction}
\label{sec:msp}
Our auxiliary task of motion similarity prediction follows the intuition that to edit a motion, the model must first identify the parts that need editing. Put differently, given the source and the edited motions, it should quantify the similarities between them. To design this auxiliary task, we consider the following two aspects: 
\begin{itemize}
\item Constructing a \textit{predictable similarity curve} (over time) between the source and the edited motions, and
\item Designing an \textit{auxiliary loss function} that synergizes with motion editing.
\end{itemize}

\subsubsection{Predictable Motion Similarity Curve}
In our work, we consider the motion similarity $\mathbf{S}^R$ between two 3D motion sequences in the space of joint rotations to leverage its scale-agnostic nature. In the rotation space, we define the raw similarity $S_{i}^{Rr}$ of one frame $i$ in $X$ as the minimum distance between this frame and any frame within a corresponding sliding window of the edited motion, as
\begin{equation}
    S_{i}^{Rr} = -\min_{\modulus{i-j}\le W} d_{r}\parens{x^i,  m^j},
    \label{eqn:sliding_window}
\end{equation}
where $W$ is the size of the sliding window, and $d_{r}$ represents a distance metric (\textit{e.g.}, Euclidean) between the source and the edited poses. The sliding window approach allows each source frame to \textit{find} its best match in the edited motion within an acceptable range of misalignment. Note that if, instead, we calculate the difference between corresponding frames with identical indices, \textit{i.e.}, $x^i$ and $m^i$, the raw similarity $S^r$ would be low for motions that are nearly identical but shifted by a few frames. Symmetrically, we define $S_{i}^{Rl}$ in the joint location space.
We balance the effects of $S^{Rl}$ and $S^{Rr}$ by empirically weighting them, as
\begin{equation}
    S_{i}^{R} = w_1{S^{Rr}_i} + w_2{S^{Rl}_i}.
\end{equation}

\begin{figure}[t]
  \centering
   \includegraphics[width=\columnwidth]{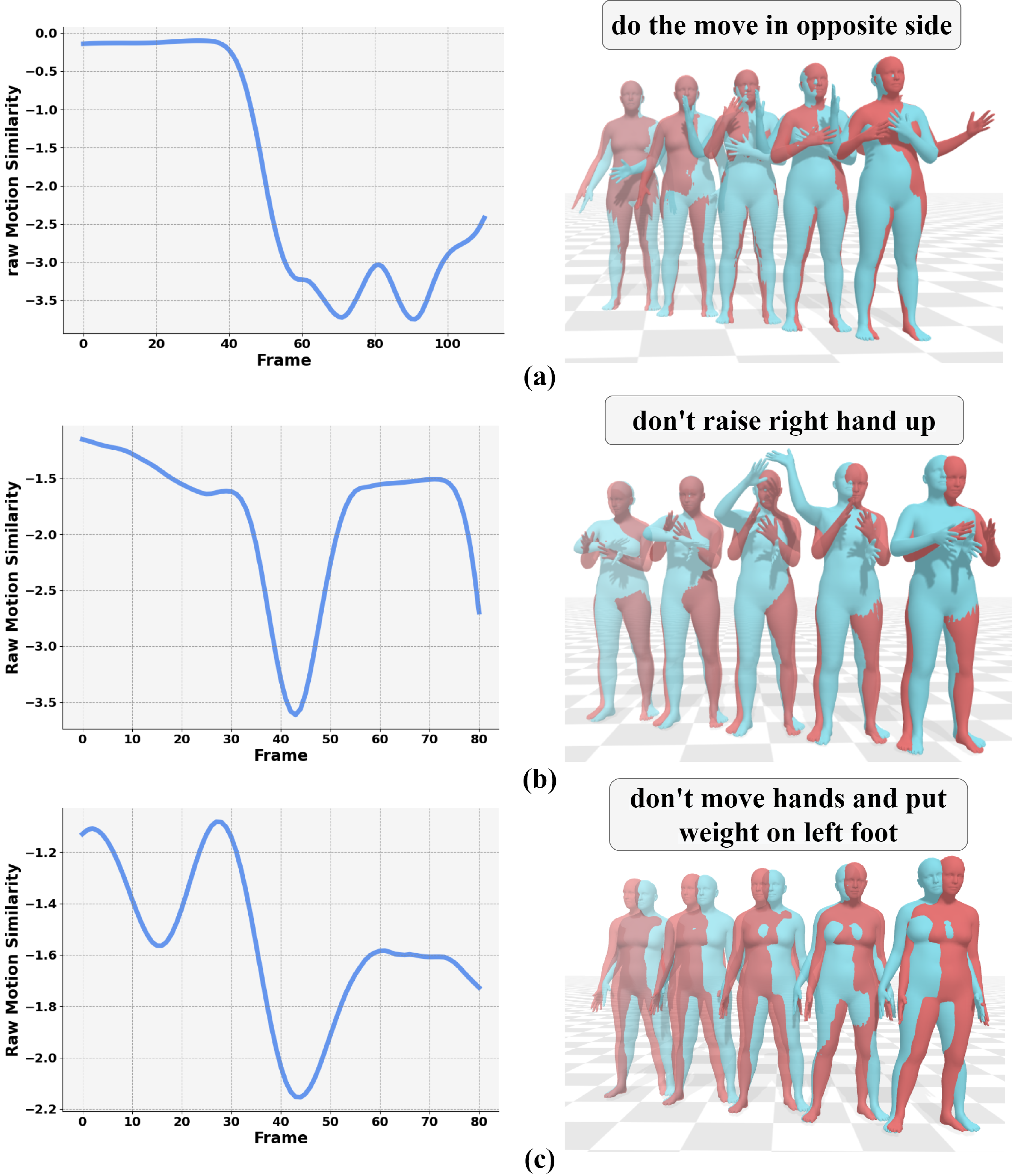}
   \caption{\textbf{Raw Motion Similarity}. We translate the global positions of sampled poses of the \textcolor{blue}{source motion} and the \textcolor{red}{edited motion} for a clear view.}
   \label{fig:sim}
\end{figure}

However, $S_{i}^{R}$ is not entirely predictable due to two issues. First, different editing instructions can result in varying scales of change. For instance, both (a) and (b) in \cref{fig:sim} represent simple edits. Yet, $S_{i}^{R}$ starts near $0$ in (a) and near $-1$ in (b). Such variations are challenging for the model to learn. Second, when the edit is fine-grained, the subtle changes make the curve appear random. For example, in \cref{fig:sim} (c), the hand movements corresponding to the text instructions ``don't move hands and put weight on left foot'' are subtle in the source motion, so the differences in the edited motion are minimal. As a result, the corresponding raw motion similarity curve lacks any identifiable patterns.

To address the first issue, we normalize all values in $S^R$ to $\bracks{0, 1}$ with min-max normalization and make the curve scale-agnostic, as
\begin{equation}
    S_{i}^N = \frac{S_{i}^{r} - \min_j S_j^R}{ \max_j S_j^R - \min_j S_j^R}.
    \label{eqn:normalization}
\end{equation}

Our sliding window (\cref{eqn:sliding_window}) and normalization (\cref{eqn:normalization}) approaches also address the second issue. 
In addition, we filter out data samples that exhibit a sufficiently large semantic gap between the text instructions and the motion similarity from our model training, as these introduce unnecessary noise in the similarity curve.
To this end, we define a motion signal-to-noise ratio (MotionSNR) based on the similarities of the top-$\kappa$ and bottom-$\kappa$ frames in a sequence, calculated as follows:
\begin{align}
    \mathbf{T}^R &= \{S_k^R \mid S_k^R \geq S_j^R &\forall S_j^R \in \mathbf{S}^R - \mathbf{T}^R \nonumber\\
    && \textrm{and} \modulus{\mathbf{T}^R} = \kappa\}, \\
    \mathbf{B}^R &= \{S_k^R \mid S_k^R \leq S_j^R &\forall S_j^R \in \mathbf{S}^R - \mathbf{B}^R \nonumber\\
    && \textrm{and} \modulus{\mathbf{B}^R} = \kappa\}, \\
    \text{MotionSNR} &= \frac{\sum_{x \in \mathbf{T}^R} x}{\sum_{x \in \mathbf{B}^R} x}. &
\end{align}
For our model training, we choose $\kappa = 5$.

Ideally, the motion for non-edited frames should remain the same as the source motion, while for edited frames, it should differ. Therefore, MotionSNR should be infinite for an ideal pair of source and edited motions. In practice, motion pairs with high MotionSNR should be easier for training, and the similarities should be easier to predict. Therefore, we filter out sequences with low MotionSNR using an empirically determined threshold.
We verify the impact of this filtering approach in \cref{subsec:quant}.

\subsubsection{Auxiliary Loss Function}
An underlying assumption of text-based motion editing is that the source motion and the text instructions should collectively predict the edited motion. By extension, we assume that the source motion and the text instructions should also collectively predict the motion similarity. Additionally, we note that motion similarity should contain sufficient information for motion editing. For instance, if the text instruction is ``use both hands to catch the ball'', motion similarity should identify which frames involve using only one hand to catch the ball and, therefore, need editing. We capture this notion of motion similarity through our auxiliary loss function. To implement this loss, instead of directly regressing the motion similarity curve, we quantize the values in $S^N$ into $K$ classes. Given motion similarity $S_i^N$, we use a quantization function $\mathcal{Q}\parens{\cdot}$ as
\begin{equation}
    \mathfrak{s}_i = \mathcal{Q}\parens{S_i^N} :=
    \begin{cases}
    0, & S_i^N < \tau_0, \\
    1, & \tau_0 \le S_i^N < \tau_1, \\
    \vdots & \vdots \\
    K-1, & S_i^N \ge \tau_{K-2},
    \end{cases}
\end{equation}
where $\mathfrak{s}_i \in \braces{0, \dots, K-1}$ provides the class label corresponding to each similarity level. We split the space $\bracks{0, 1}$ into $K$ equal-length intervals for the similarity levels.

For each frame $i$, assuming the network outputs logits $\mathbf{z}_i = \bracks{z_{i, 0}, \dots, z_{i, K-1}}$, we convert logits to probabilities using the softmax function
\begin{equation}
    p_{i, k} = \frac{\exp\parens{z_{i, k}}}{\sum_{l=0}^{K-1} \exp\parens{z_{i, l}}} \quad \forall k \in \braces{0, \dots, K-1}.
\end{equation}

We can then write the cross-entropy loss for the classification task, which gives our auxiliary loss function, as
\begin{equation}
    \mathcal{L}_{\text{aux}} = -\frac{1}{F} \sum_{i=0}^{F-1} \log p_{i,\mathfrak{s}_i}
    \label{eqn:loss_aux}
\end{equation}

Quantizing the similarity values provides the following benefit. For a given pairing of source motion, text instructions, and edited motion, the edited motion is only one of all feasible edited motions given just the source motion and the text instructions. Therefore, regression-based learning confines the model's ability to see only the provided edited motions as feasible. Quantization, by contrast, allows the model to focus on the coarser similarity levels rather than perfect frame-level similarity, striking a balance between optimizing the auxiliary and the editing losses and also enabling the model to generalize to a wider range of text instructions and motion variations than what is available from the dataset. We further validate this design choice experimentally in \cref{tab:aux}. 

Lastly, combining the editing and the auxiliary losses (\cref{eqn:loss_editing,eqn:loss_aux}), we write the total training loss as
\begin{equation}
   \mathcal{L} =  \mathcal{L}_{\text{aux}} + \mathcal{L}_\text{e}.
\end{equation}

\subsection{Motion Diffusion Transformer}
\label{sec:dit}
We combine diffusion-based motion editing and the auxiliary task of motion similarity prediction into our Motion Diffusion Transformer model \cref{fig:pipeline}. Our condition transformer follows the traditional transformer encoder architecture. It takes in the source motions and the CLIP features~\cite{radford2021learning} of the text instructions, and outputs enhanced text features and motion features that contain sufficient information for the desired edited motions. This enhancement of the text and motion features via their intermixing is a direct consequence of using the auxiliary loss function (\cref{eqn:loss_aux}). The diffusion transformer takes in the enhanced motion features concatenated with the noised edited motions. We also inject the enhanced text features into the diffusion transformer through an AdaLN-Zero layer, alongside the diffusion timestep $t$. We use the editing loss (\cref{eqn:loss_editing}) for the model to learn the edited motions at the output of the diffusion transformer, and train it jointly with the auxiliary task. We adopt the standard DDPM diffusion paradigm~\cite{ddpm1} for both training and inference.

Our design ensures that the inputs to the auxiliary task are decoupled from the motion editing task. This is desirable since any changes in the noised edited motion (part of the diffusion transformer's inputs) should not affect the model's learning of the auxiliary task. Conversely, our design also ensures that the motion editing task fully leverages the enhanced motion and text features learned from the auxiliary task by directly using them as inputs for the denoising diffusion process.
\section{Experiments and Results}
\label{sec:exp}
We detail our experimental setup, including dataset, metrics, baselines, and implementation details, and elaborate on the results.

\begin{figure*}[!htbp]
    \centering
    \includegraphics[width=0.95\textwidth]{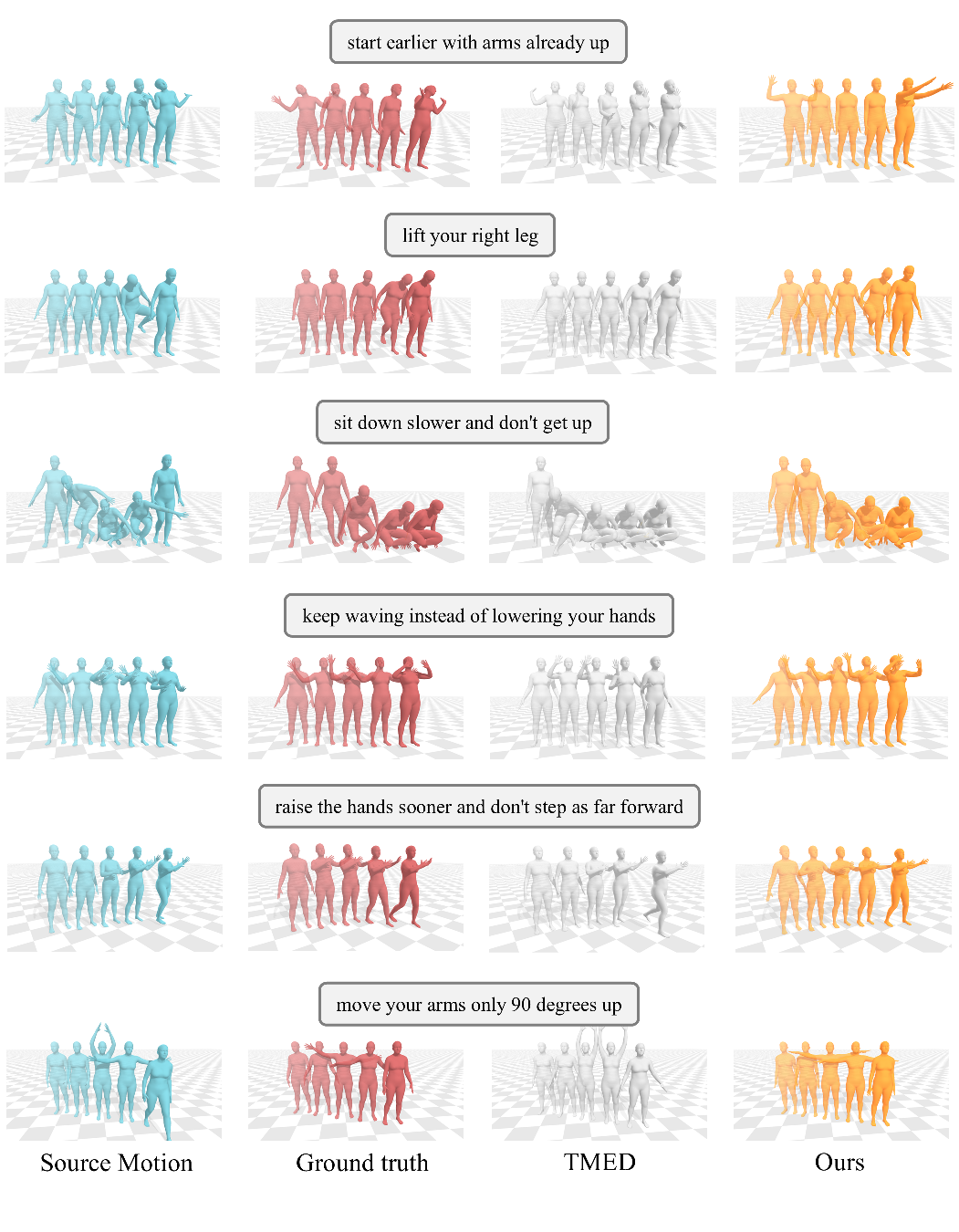}
    \caption{\textbf{Qualitative Results.} We compare our method with TMED~\cite{athanasiou2024motionfix}. Our method outperforms TMED in terms of both fidelity and alignment with source motion and text instructions.}
    \label{fig:qual}
\end{figure*}

\begin{table*}[t]
    \centering
    \caption{\textbf{Comparison of Text-Based Motion Editing on the MotionFix~\cite{athanasiou2024motionfix} Dataset.} 
    Our method outperforms all other baselines on generated-to-target retrieval and M-score. We report R@1, R@2 and R@3 as percentages.
     $\uparrow$ indicates higher values are better, $\downarrow$ indicates lower values are better, bold indicates \textbf{best}, and underline indicates \underline{second best}.}
    \label{tab:main}
    \begin{tabular}{lccccL{0.01cm}ccccc}
        \toprule
        \multirow{2}{*}{Methods} & \multicolumn{4}{c}{Generated-to-Target (Batch)} && \multicolumn{4}{c}{Generated-to-Target (Test Set)} & \multirow{2}{*}{M-score$\uparrow$} \\
        \cmidrule{2-5}\cmidrule{7-10}
         & R@1$\uparrow$ & R@2$\uparrow$ & R@3$\uparrow$ & AvgR$\downarrow$ && R@1$\uparrow$ & R@2$\uparrow$ & R@3$\uparrow$ & AvgR$\downarrow$ \\
        \midrule
        \cellcolor{light-gray}{Ground Truth} & \cellcolor{light-gray}{$100.0$} & \cellcolor{light-gray}{$100.0$} & \cellcolor{light-gray}{$100.0$} & \cellcolor{light-gray}{$1.00$} & \cellcolor{light-gray}{ } & \cellcolor{light-gray}{$64.36$} & \cellcolor{light-gray}{$88.75$} & \cellcolor{light-gray}{$95.56$} & \cellcolor{light-gray}{$1.74$} & \cellcolor{light-gray}{$-3.175$} \\
        \midrule
        MDM~\cite{mdm} & $4.03$ & $7.56$ & $10.48$ & $15.55$ && $0.10$ & $0.10$ & $0.10$ & $-$ & $-$ \\
        MDM-BP~\cite{athanasiou2024motionfix} & $39.10$ & $50.09$ & $54.84$ & $6.46$ && $8.69$ & $14.71$ & $18.36$ & $180.99$ & $-$ \\
        TMED~\cite{athanasiou2024motionfix} & $62.90$ & $76.51$ & $83.06$ & $2.71$ && $14.51$ & $21.72$ & $28.73$ & $56.63$ & $-3.512$  \\
        \midrule
        Ours & $\mathbf{70.62}$ & $\mathbf{82.92}$ & $\mathbf{88.12}$ & $\mathbf{2.38}$ && $\mathbf{25.49}$ & $\mathbf{39.33}$ & $\mathbf{49.21}$ & $\mathbf{23.49}$ & $\mathbf{-3.210}$ \\
        Ours (w/o aux loss) & $\underline{67.21}$ & $78.75$ & $\underline{86.33}$ & $2.49$ && $23.32$ & 
        $\underline{36.96}$ & $\underline{46.44}$ & $26.02$ & $-3.422$ \\
        Ours (w/o filtering) & $66.88$ & $\underline{80.42}$ & $85.83$ & $\underline{2.46}$ && $\underline{23.73}$ & $35.57$ & $45.36$ & $\underline{24.94}$ & $\underline{-3.418}$ \\
        \bottomrule
    \end{tabular}
\end{table*}

\subsection{Setup}
\paragraph{Data.}
The MotionFix dataset~\cite{athanasiou2024motionfix} is a language-based 3D motion editing dataset. Each data sample is a triplet consisting of a source motion, a target motion, and a text instruction. To ensure quality, semantically similar motion pairs are selected from motion capture (MoCap) data using the TMR motion embedding space~\cite{petrovich2023tmr}. Annotators then manually described the differences in a concise manner. MotionFix provides a total of $6{,}730$ annotated triplets split into train, validation, and test sets.
\paragraph{Evaluation Metrics.}
For text-based human motion editing, we need to evaluate both the \textit{semantic alignment} of the edited motions with the text instructions and the source motions, and the \textit{fidelity} of the edited motions.
Following the approach of MotionFix~\cite{athanasiou2024motionfix}, we evaluate \textit{alignment} based on motion-to-motion retrieval. We utilize pre-trained TMR~\cite{petrovich2023tmr} as a feature extractor and compute the retrieval accuracy of the generated motion in a fixed-size batch of motions. We report the top-1, top-2, and top-3 accuracies (R@1, R@2, and R@3) both in batch sizes of 32 and in the full test set. We also report the average rank for comparison. To evaluate \textit{fidelity}, we utilize the MotionCritic score (M-score)~\cite{wang2024aligning}. M-score has been shown to align with human perceptual preferences, thus serving as a reasonable metric for motion realism. To compute M-scores, we use the authors' pre-trained model and transform our generated motions to match the pre-trained model's input.

\paragraph{Baselines.}
We compare our method with three baselines, MDM~\cite{mdm}, MDM-BP~\cite{athanasiou2024motionfix}, and TMED~\cite{athanasiou2024motionfix}.
\begin{itemize}
    \item MDM utilizes the pre-trained human motion diffusion model. The edit instruction is taken as text input.
    \item MDM-BP enhances MDM by introducing body-part detection using a GPT-based annotation system~\cite{achiam2023gpt}. The method identifies body parts not mentioned in the text instructions, allowing the model to mask these parts and retain their source motion features.
    \item TMED is a conditional diffusion model specifically trained on the MotionFix dataset using motion-motion-text triplets. Unlike the baselines trained only on text-motion pairs, TMED leverages both the source motions and the text instruction as conditions to generate the desired edited motions.
\end{itemize}
Each of our baselines serves a specific role in evaluations. MDM represents the editing power of the diffusion model without the knowledge of source motions. MDM-BP knows both the source motions and the text instructions but is not trained on text-based motion editing datasets. TMED takes both source motions and text as input and is trained on the dataset, which is comparable to our approach, albeit with a different model architecture.

\paragraph{Implementation Details.}
We fix the DDPM~\cite{ddpm1} diffusion step to 300 and use the cosine noise scheduler. Following \cite{athanasiou2024motionfix} and \cite{brooks2023instructpix2pix}, we adapt two-way conditioning with a guidance scale of $2$ for both the text and the source motion. We utilize pre-trained CLIP~\cite{radford2021learning} from ViT-B/32 to encode the text. Our condition and diffusion transformers consist of $4$ and $8$ transformer encoder layers, respectively, $8$ attention heads, and latent dimensions of $512$. We train using a batch size of $128$, and the AdamW optimizer~\cite{loshchilov2017decoupled} with a learning rate of $10^{-4}$.
We train our model for $1{,}500$ epochs, which takes around a day on a single A100 GPU.

\subsection{Quantitative Comparisons}
\label{subsec:quant}
We report quantitative comparisons in \cref{tab:main}. Our model outperforms all baselines significantly in both motion-editing alignment and fidelity. MDM shows the weakest performance, aligning with the intuition that text instructions by themselves do not provide sufficient information to generate source-aligned edited motions. With GPT assistance in identifying edited body parts, MDM-BP shows substantial improvements over MDM. TEMD considers the influence of text instructions on the source motion and further outperforms MDM-BP. Our Motion Diffusion Transformer, together with the auxiliary loss for the learning of motion similarities, results in state-of-the-art performance. We attribute the progressive performance improvements to the knowledge of the source motion, effective dataset utilization, a specially designed network architecture, and an appropriate auxiliary loss function. We also evaluate our variant without filtering out noisy motion similarity curves for the auxiliary loss (w/o filtering -- \cref{tab:main}, last row). The performance downgrade implies that noisy data for motion similarity prediction negatively affects the training.

\subsection{Qualitative Results}
We show visual results compared with TMED~\cite{athanasiou2024motionfix} in \cref{fig:qual}. Our method improves on various generative artifacts of TMED, such as self-penetrations (row 1), failure to follow instruction nuances (row 2 -- not generating motions with the right leg lifted, and row 3 -- not generating the sitting down motion to be slower), and failure to maintain the effects of specific instructions (row 4 -- lowering the hands instead of continuing to wave, row 5 -- unexpectedly stepping forward, and row 6 -- moving the arms beyond 90 degrees). These examples also illustrate the realism of our generated edited motions.

\subsection{Ablation Study}
We perform ablations on both the Motion Diffusion Transformer and motion similarity prediction.

\subsubsection{Motion Diffusion Transformer}

\begin{table}[t]
\centering
\caption{\textbf{Performances of \ours with Different Combinations of Condition Features.} In the ``no text feature'' setting, the input to the condition transformer is the raw text. Using enhanced features for both text and motion leads to the best overall performance for motion realism.}
\label{tab:arch}
\resizebox{\columnwidth}{!}
{
\begin{tabular}{llcccc}
    \toprule
    \multirow{2}{*}{Text Feat.} & \multirow{2}{*}{Motion Feat.} & \multicolumn{3}{c}{Generated-to-Target (Batch)} & \multirow{2}{*}{M-Score$\uparrow$}\\
    \cmidrule{3-5}
    & & R@1$\uparrow$ & R@2$\uparrow$ & R@3$\uparrow$ &  \\
    \midrule
    Raw & Enhanced & $70.42$ & $81.25$ & $86.67$ & $-3.471$ \\
    Raw & Raw & $67.21$ & $78.75$ & $86.33$ & $-3.422$ \\
    Enhanced & Enhanced & $\underline{70.62}$ & $\underline{82.92}$ & $\underline{88.12}$ & $\mathbf{-3.210}$ \\
    Enhanced & Raw & $69.08$ & $79.75$ & $86.59$ & $-3.450$ \\
    Enhanced & No & $66.46$ & $81.25$ & $84.38$ & $-3.611$ \\
    No & Enhanced & $\mathbf{71.04}$ & $\mathbf{83.96}$ & $\mathbf{89.58}$ & $\underline{-3.318}$ \\
    \bottomrule
    \end{tabular}
}
\end{table}

We report comparisons with variations of our architecture in \cref{tab:arch}. The variant with both enhanced text and enhanced source motion features outperforms most of the other variants. It falls marginally behind on motion alignment and somewhat ahead on motion fidelity compared to using enhanced motion features with the raw text, indicating the strong influence of the source motion on aligning the edited motion but not having much influence on its fidelity. This outcome also suggests that the joint enhancement of both text and motion features provides a better balance between alignment and realism (fidelity). With these findings, we validate the effectiveness of enhanced features produced by the condition transformer and motion similarity prediction.

\subsubsection{Motion Similarity Prediction}
\begin{table}[t]
\centering
\caption{\textbf{Performances of \ours with Different Auxiliary Losses.} Model performance improves when using a classification loss over a regression loss for the auxiliary task, indicating the better learning capacity enabled by classification. However, performance also decreases when increasing the number of classes beyond three, which brings the classes too close to each other.}
\label{tab:aux}
\resizebox{\columnwidth}{!}
{
    \begin{tabular}{lccccc}
    \toprule
    \multirow{2}{*}{Loss} & \multirow{2}{*}{\#classes} & \multicolumn{3}{c}{Generated-to-Target (Batch)} & \multirow{2}{*}{M-Score$\uparrow$}\\
    \cmidrule{3-5}
    & & R@1$\uparrow$ & R@2$\uparrow$ & R@3$\uparrow$ &  \\
    \midrule
    Reg. & $\infty$ & $68.96$ & $82.71$ & $87.08$ & $-3.750$ \\
    Cls. & $3$ & $\mathbf{70.62}$ & $\mathbf{82.92}$ & $\mathbf{88.12}$ & $\underline{-3.210}$ \\
    Cls. & $5$ & $67.08$ & $80.21$ & $86.46$ & $\mathbf{-3.186}$ \\
    Cls. & $9$ & $66.88$ & $82.29$ & $86.88$ & $-3.722$ \\
    \bottomrule
    \end{tabular}
}
\end{table}

We report the results of ablating the auxiliary loss in ~\cref{tab:aux}. We find that quantizing the motion similarity values into three classes yields the best results, implying that the auxiliary task should be simple enough to not overpower the main editing task.
Further, a regression-based auxiliary loss reduces the motion fidelity given the overfitting tendencies when attempting to precisely predict the motion similarity curves.

\section{Conclusion}
\label{sec:conclusion}
We have presented \ours, a state-of-the-art solution for text-based human motion editing. Our method utilizes the Motion Diffusion Transformer architecture to inject the source motion and the text instructions into a diffusion model for predicting the edited motions. We also leverage the novel auxiliary task of motion similarity prediction to further enhance the interplay of text and source motion features in the context of motion editing.
Extensive experiments demonstrate that \ours outperforms prior works in both alignment and realism of the edited motions. Through detailed ablation studies, we verify the effects of the core components of our method. We hope our work can push the limits of text-based human motion editing and inspire future research.

\section*{Acknowledgements}
We would like to thank Nikos Athanasiou for helpful discussions and assistance with the evaluation.



{
    \small
    \bibliographystyle{ieeenat_fullname}
    \bibliography{main}
}
\clearpage
\maketitlesupplementary

\renewcommand{\thesection}{\Alph{section}}
\setcounter{section}{0}
\counterwithin{equation}{section}
\counterwithin{figure}{section}
\counterwithin{table}{section}




\section{Perceptual Study}\label{sec:user_study}

\begin{figure*}[t]
    \centering
    \begin{subfigure}[b]{0.45\textwidth} 
        \includegraphics[clip, trim=0 3.95in 0 0, width=\textwidth]{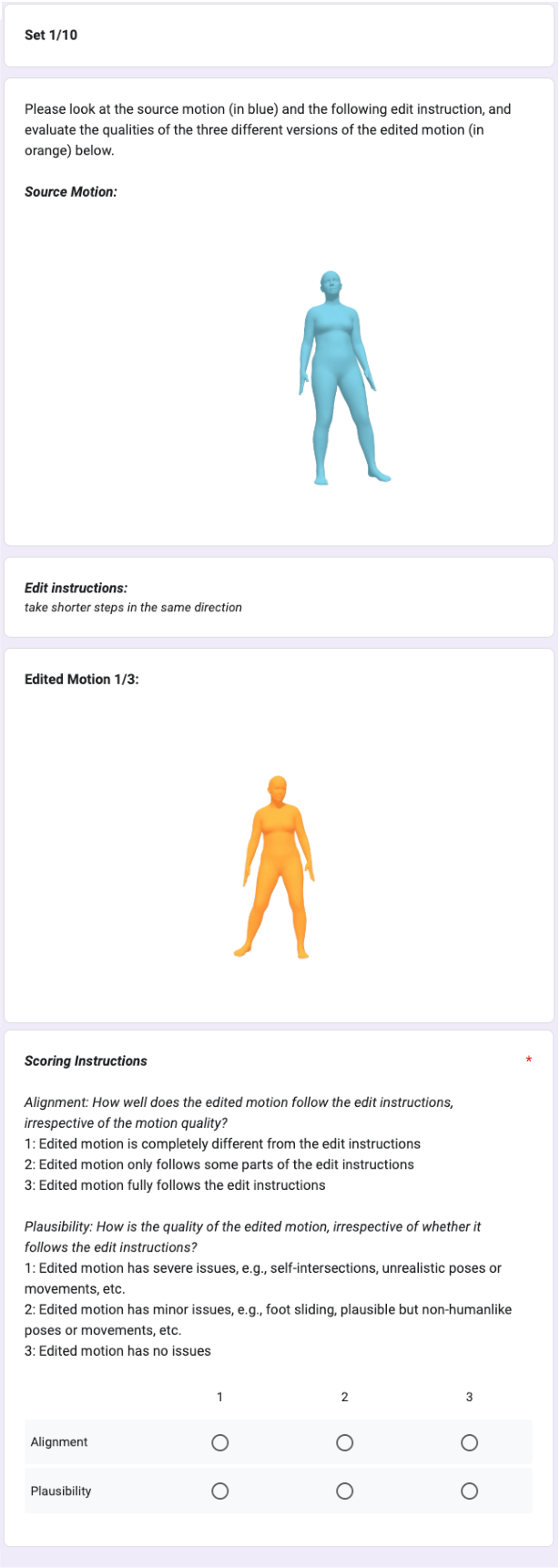}
        \caption{Upper Part}
        \label{fig:user_study_layout_upper}
    \end{subfigure}
    \hfill 
    \begin{subfigure}[b]{0.45\textwidth} 
        \includegraphics[clip, trim=0 0 0 7.5in, width=\textwidth]{images/user_study_layout.eps}
        \caption{Lower Part}
        \label{fig:user_study_layout_lower}
    \end{subfigure}
    \caption{\textbf{Perceptual Study Layout.} \textit{(Upper Part)} We show an example of our study layout with one source motion, one edit instruction, and one edited motion. \textit{(Lower Part)} We show the scoring instructions and scoring area for all the samples in the perceptual study.}
    \label{fig:user_study_layout}
\end{figure*}


In addition to our quantitative and qualitative comparisons, we have also conducted a perceptual study with human participants to understand how human observers view the quality of motions edited by \ours, particularly in comparison to the best available baseline TMED~\cite{athanasiou2024motionfix} and the corresponding ground truth edited motions.

\subsection{Setup}\label{subsec:study_setup}
Each participant in our study observed 10 sample sets corresponding to 10 randomly selected pairings of source motions and textual edit instructions. Within each sample set, a participant observed the source motion, the text instructions, and three versions of the edited motion --- one each for the ground truth, \ours, and TMED. We put these three edited motions in a random order within each sample set to avoid any positional biases in the study. We asked the participants to respond on two metrics:
\begin{itemize}
    \item \textit{Alignment of edited motions with text: How well does the edited motion follow the edit instructions, irrespective of the motion quality?} On this metric, the participants responded in a 3-point Likert Scale with the following descriptions: ``Edited motion is completely different from the edit instructions'' (1), ``Edited motion only follows some parts of the edit instructions'' (2), and ``Edited motion fully follows the edit instructions'' (3),
    \item \textit{Plausibility of edited motions: How is the quality of the edited motion, irrespective of whether it follows the edit instructions?} On this metric, the participants responded in a 3-point Likert Scale with the following descriptions: ``Edited motion has severe issues, \textit{e.g.}, self-intersections, unrealistic poses or movements, \textit{etc.} (1), ``Edited motion has minor issues, \textit{e.g.}, foot sliding, plausible but non-humanlike poses or movements, \textit{etc.}'' (2), and ``Edited motion has no issues'' (3).
\end{itemize}

We show an example layout with one source motion, one edit instruction, and one edited motion in \cref{fig:user_study_layout_upper} and the scoring instructions and scoring area for the participants in \cref{fig:user_study_layout_lower}.

\subsection{Results}\label{subsec:study_results}

\begin{figure*}[t]
\centering

\mpage{0.48}{\includegraphics[width=\linewidth, trim=0 1.2cm 0 0, clip]{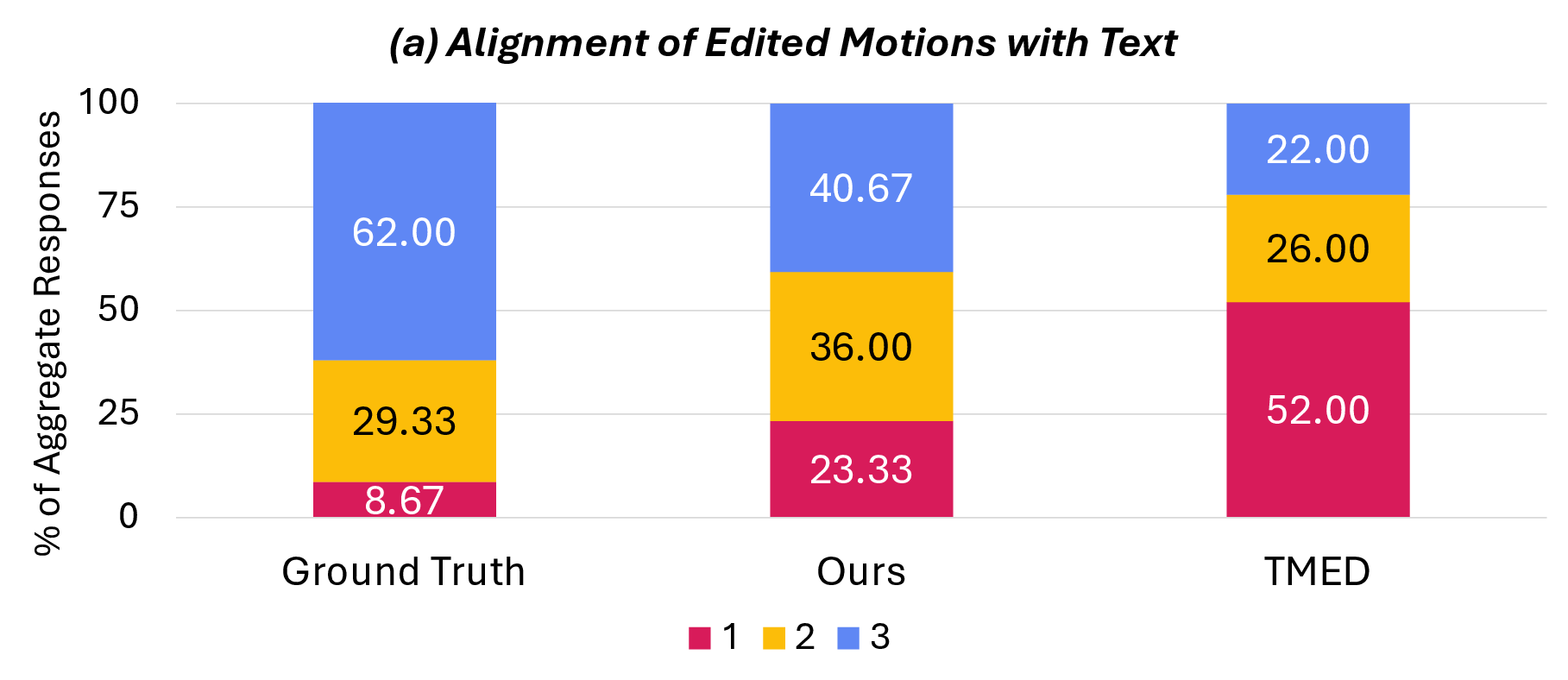}}\hfill
\mpage{0.48}{\includegraphics[width=\linewidth, trim=0 1.2cm 0 0, clip]{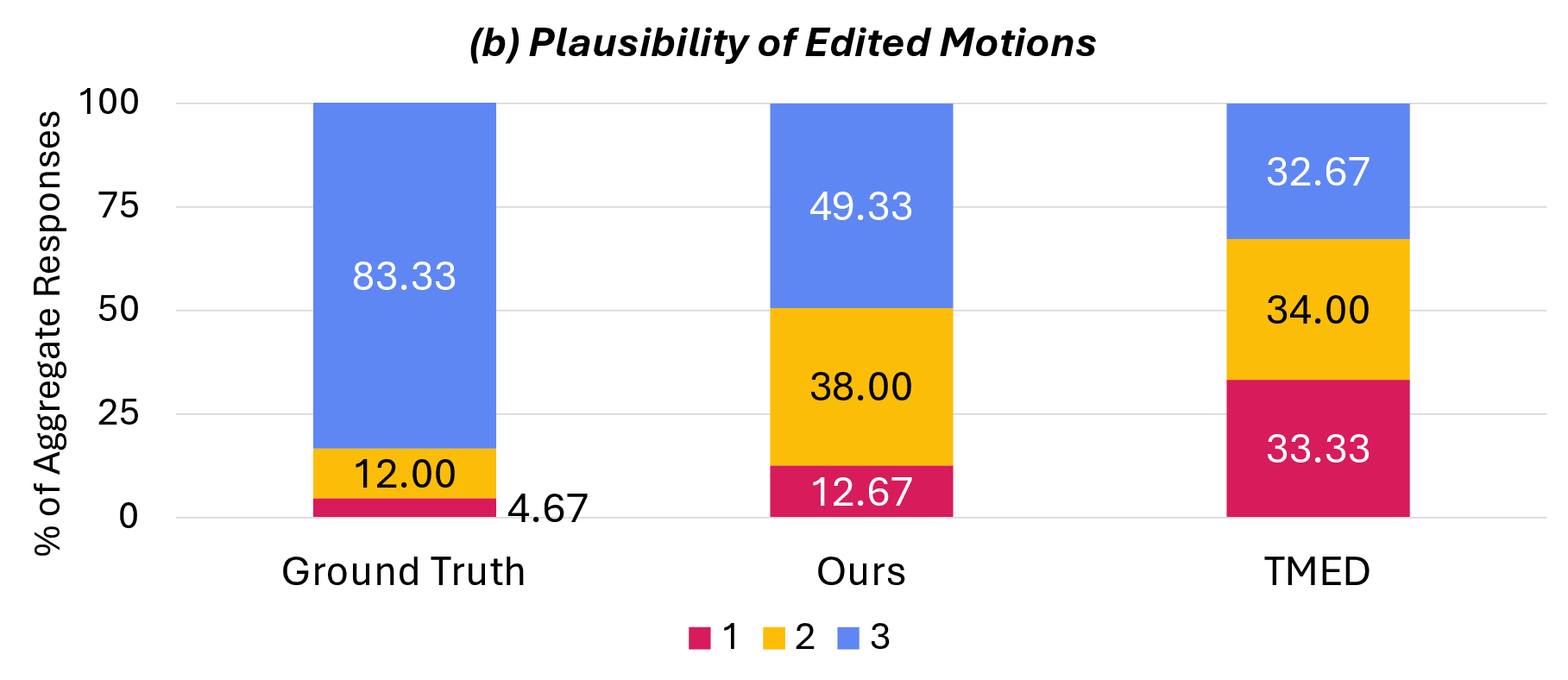}}\\
\mpage{1.0}{\includegraphics[width=.5\linewidth, trim=0 0cm 0cm 12cm, clip]{images/user_study_scores_alignment.eps}}\\

\caption{\textbf{Perceptual Evaluation Score Distributions.} We show the distributions of aggregate responses from participants on the three versions of edited motions --- Ground Truth, \ours, and TMED~\cite{athanasiou2024motionfix} --- on the two metrics of \textit{Alignment} and \textit{Plausibility}. We observe that participants have marked 3 for \ours about $40\%$ to $50\%$ more times than TMED across the two metrics.}
\label{fig:user_study_scores}
\end{figure*}

A total of 15 participants, consisting of students and staff from a University campus, responded to our perceptual study, leading to a total of 150 responses across the 10 sample sets. We show the distribution of participant scores aggregated over all the responses in \cref{fig:user_study_scores}, and report the mean statistics in \cref{tab:user_study_stats}. We observe that \ours is consistently scored higher than TMED~\cite{athanasiou2024motionfix}, outperforming it by $40\%$ to $50\%$ on score distributions and by an absolute $0.3$ to $0.5$ points on the mean score on a 3-point Likert Scale across the two metrics.

\begin{table}[t]
\centering
\caption{\textbf{Perceptual Evaluation Mean Statistics.} We report the mean scores achieved by all three candidates in the perceptual study, averaging the aggregated responses across all the participants and sample sets. \ours achieves scores that are $0.3$ to $0.5$ points higher than TMED on a 3-point Likert Scale.}
\label{tab:user_study_stats}
\begin{tabular}{llc}
\toprule
Metric & Candidate & Mean Score $\uparrow$\\
\midrule
\textit{Alignment} & \cellcolor{light-gray}{Ground Truth} & \cellcolor{light-gray}{$2.53 \pm 0.65$} \\
\cmidrule{2-3}
& \ours (ours) & $\mathbf{2.17 \pm 0.78}$ \\
& TMED~\cite{athanasiou2024motionfix} & $1.70 \pm 0.81$ \\
\midrule
\textit{Plausibility} & \cellcolor{light-gray}{Ground Truth} & \cellcolor{light-gray}{$2.79 \pm 0.51$} \\
\cmidrule{2-3}
& \ours (ours) & $\mathbf{2.37 \pm 0.70}$ \\
& TMED~\cite{athanasiou2024motionfix} & $1.99 \pm 0.81$ \\
\bottomrule
\end{tabular}
\end{table}

\section{Additional Quantitative Evaluations} 
For the preservation of the source motion, we follow TMED and use generated-to-source retrieval to measure motion preservation. \cref{tab:add} shows the comparable performance of our method to TMED. For other fidelity-related metrics, since TMED does not provide the implementation of FID and L2 distance, we implement our own for a fair comparison. We see that our method outperforms TMED. 

\begin{table}[t]
\centering
\caption{\textbf{Additional Performance and Efficiency Evaluations.} Compared to TMED, we report comparable generated-to-source accuracy measures and lower L@ distance and FID measures.}
\label{tab:add}
\resizebox{\columnwidth}{!}
{
    \begin{tabular}{lccccc}
    \toprule
    Method & \multicolumn{3}{c}{Generated-to-Source (Batch)} & L2 Dist. & FID \\
    \cmidrule{2-4}
    & R@1$\uparrow$ & R@2$\uparrow$ & R@3$\uparrow$ & (m)$\downarrow$ & $\downarrow$ \\
    \midrule
    \cellcolor{light-gray}{GT} & \cellcolor{light-gray}{$74.01$} & \cellcolor{light-gray}{$84.52$} & \cellcolor{light-gray}{$89.91$} & \cellcolor{light-gray}{$-$} & \cellcolor{light-gray}{$-$} \\
    TMED & $71.77$ & $\mathbf{84.07}$ & $\mathbf{89.52}$ & $0.278$ & $0.167$ \\
    Ours & $\mathbf{72.71}$ & $83.54$ & $87.50$ & $\mathbf{0.253}$ & $\mathbf{0.110}$ \\
    \bottomrule
    \end{tabular}
}
\end{table}


\section{Additional Qualitative Results} 

\begin{figure*}[!htbp]
    \centering
    \includegraphics[width=0.95\textwidth]{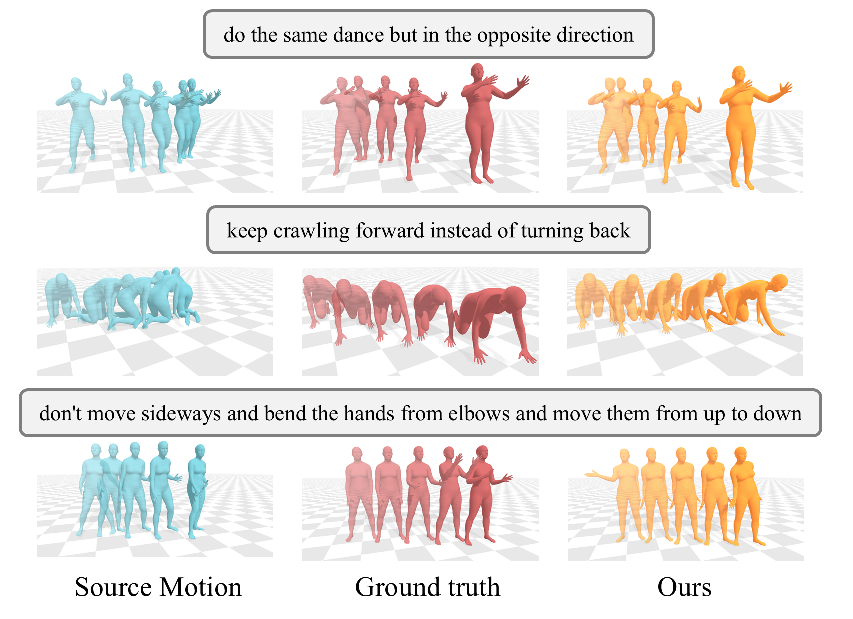}
    \caption{\textbf{More Qualitative Results.}}
    \label{supfig:qual}
\end{figure*}

As shown in Fig.~\ref{supfig:qual}, our method can edit dance and crawling motions. It also successfully follows complex edit instructions.
\end{document}